\documentclass[letterpaper, 10 pt, conference]{ieeeconf}  

\IEEEoverridecommandlockouts                              

\usepackage{graphics} 
\usepackage{epsfig} 
\usepackage{mathptmx} 
\usepackage{times} 
\usepackage{amsmath} 
\usepackage{amssymb}  

\usepackage[top=52pt,left=48pt,right=48pt,bottom=43pt]{geometry}
\usepackage{setspace}
\usepackage{float}
\usepackage{stmaryrd}

\usepackage{bm}
\usepackage[short]{optidef}
\usepackage{algorithm}
\usepackage[noend]{algpseudocode}

\usepackage{diagbox}

\usepackage{tikzscale}
\usepackage{pgfplots}
\pgfplotsset{compat=1.9, legend style={font=\footnotesize}}
\usepgfplotslibrary{groupplots}

\usepackage{hyperref}
\hypersetup{
    colorlinks=true,
    linkcolor=blue,
    filecolor=magenta,      
    urlcolor=black,
}

\newlength{\textfloatsepsave} 
\title{\LARGE \bf
\vspace{-5.5mm}
LUCIDGames: onLine UnsCented Inverse Dynamic Games for \\
Adaptive Trajectory Prediction and Planning
\vspace{-1mm}
}

\author{Simon Le Cleac'h$^{1}$, Mac Schwager$^{2}$ and Zachary Manchester$^{3}$%
\thanks{
    This work was supported in part by NSF NRI grant 1830402 and DARPA YFA grant D18AP00064.  Toyota Research Institute (``TRI'') provided funds to assist the authors with their research, but this article solely reflects the opinions and conclusions of its authors and not TRI or any other Toyota entity.
}
\thanks{
    $^{1}$Simon Le Cleac'h is with the Department of Mechanical Engineering, 
    Stanford University, California, USA
    {\tt\small simonlc@stanford.edu}
}%
\thanks{
    $^{2}$Mac Schwager is with the Department of Aeronautics \& Astronautics, 
    Stanford University, California, USA
    {\tt\small schwager@stanford.edu}
}
\thanks{
    $^{3}$Zachary Manchester is with the Robotics Institute, 
    Carnegie Mellon University, Pennsylvania, USA
    {\tt\small zacm@cmu.edu}}%
\vspace{-1mm}
}

\begin{document}

\maketitle
\thispagestyle{empty}
\pagestyle{empty}

\begin{abstract}
    Existing game-theoretic planning methods assume that the robot knows the objective functions of the other agents \emph{a priori} while, in practical scenarios, this is rarely the case. This paper introduces LUCIDGames, an inverse optimal control algorithm that is able to estimate the other agents' objective functions in real time, and incorporate those estimates online into a receding-horizon game-theoretic planner. LUCIDGames solves the inverse optimal control problem by recasting it in a recursive parameter-estimation framework. LUCIDGames uses an unscented Kalman filter (UKF) to iteratively update a Bayesian estimate of the other agents' cost function parameters, improving that estimate online as more data is gathered from the other agents' observed trajectories. The planner then takes account of the uncertainty in the Bayesian parameter estimates of other agents by planning a trajectory for the robot subject to uncertainty ellipse constraints. The algorithm assumes no explicit communication or coordination between the robot and the other agents in the environment. An MPC implementation of LUCIDGames demonstrates real-time performance on complex autonomous driving scenarios with an update frequency of 40 Hz. Empirical results demonstrate that LUCIDGames improves the robot's performance over existing game-theoretic and traditional MPC planning approaches. Our implementation of LUCIDGames is available at \url{https://github.com/RoboticExplorationLab/LUCIDGames.jl}. 

\end{abstract}

\vspace{-1mm}
\section{Introduction}
\vspace{-1mm}
    Planning trajectories for a robot that interacts with other agents is challenging, as it requires prediction of the reactive behaviors of the other agents, in addition to planning for the robot itself. Classical approaches in the literature decouple the prediction and planning tasks. Usually, predicted trajectories of the other agents are computed first and provided as input for the robot planning module, which considers them as immutable obstacles. This formulation ignores the influence of the robot's decisions on the other agents' behaviors.  Moreover, it can lead to the ``frozen robot'' problem, where no safe path to the goal can be found by the planner \cite{Trautman2010} because of the false assumption that other agents will not yield or deviate from their predicted trajectory in response to the robot. Preserving the coupling between prediction and planning is thus key to producing richer interactive behavior for a robot acting among other agents. 
    
    \begin{figure}[t]
    \vspace{+1mm}
        \centering
        \includegraphics[width=8.60cm]{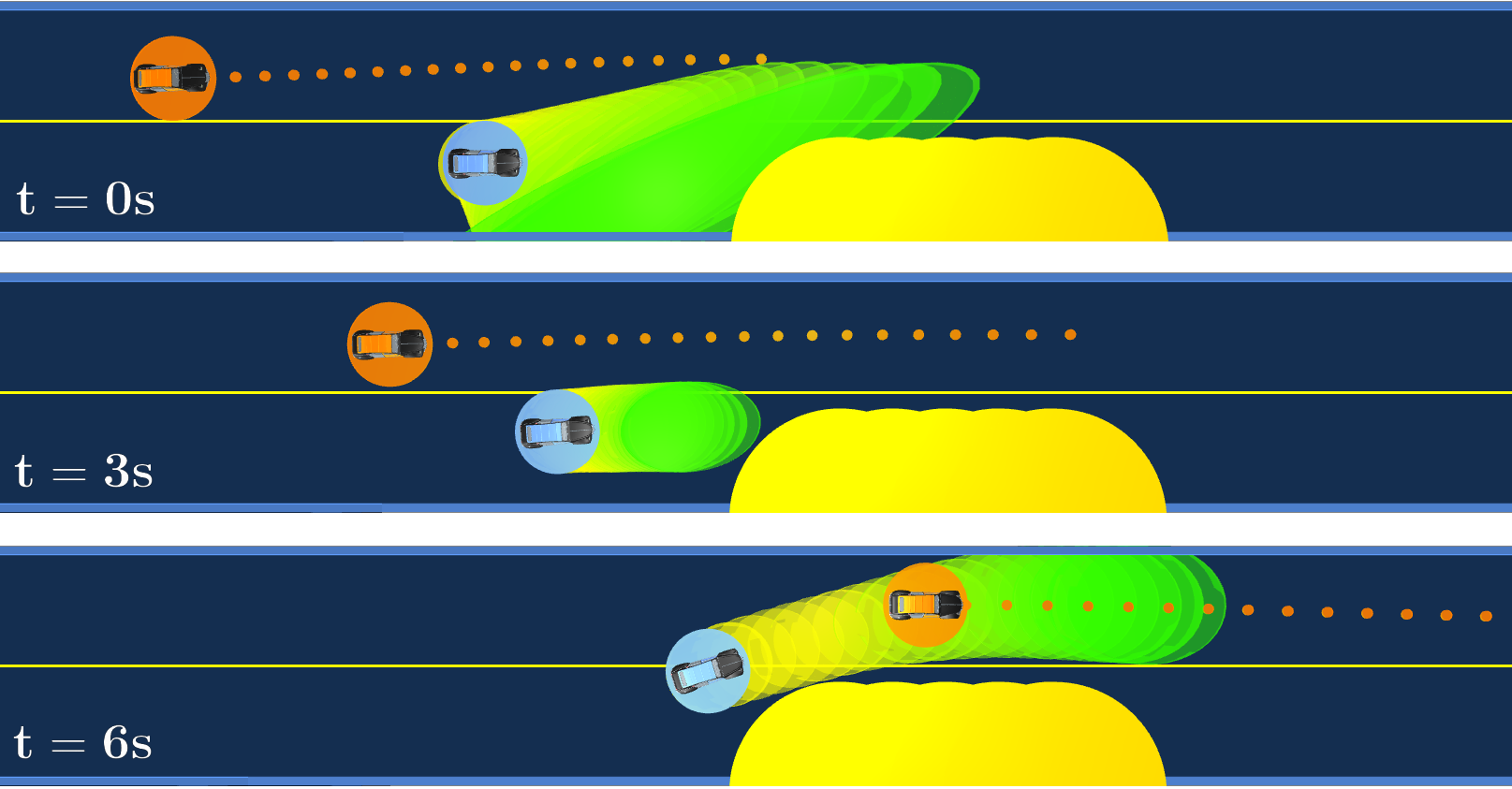}\hfill
        \caption{We present three roadway visualizations of a scenario where the autonomous vehicle (AV) in orange and a human-driven car (blue) have to overtake a large obstacle (yellow) on the bottom lane. The AV (orange) follows the robust version of LUCIDGames. At the start, the AV slows down to avoid the uncertainty-based collision avoidance zone (green), which comprises two possibilities: either the human cuts in front of the AV, or the human yields to let the AV go first. Then, by observing the human's behavior, the AV better estimates its objective and narrows down the collision avoidance zone to the first option. Finally, the AV proceeds to overtaking the obstacle before the human. The AV's planned trajectory is represented by orange dots.}
        \label{fig:ellipse_planning}
        \vspace{-7mm}
    \end{figure}

    Several recent works have used the theory of dynamic games to  capture the coupled interaction among a robot and other agents, particularly in the context of autonomous driving \cite{LeCleach2020a,Fridovich-Keil2020}.  However, these works rely on the strong assumption that the robot has full knowledge of the other agents' objective functions. In many applications, the robot only has access to a coarse estimate of these objective functions. For instance, in a crowd navigation problem, the robot might know the preferred walking speed of humans. In a ramp-merging scenario, the autonomous car might be aware of the desired distance drivers usually keep between themselves. These coarse estimates of the other agents' objectives can be obtained using real data like the NGSIM driving dataset \cite{Colyar2007} or the ETH dataset \cite{Pellegrini2009} for interacting pedestrians. Inverse reinforcement learning (IRL) approaches typically learn a general objective function to suit a large batch of demonstrations from multiple agents \cite{Ng2000, Ziebart2008, Waugh2013}. On the contrary, our goal is to accurately estimate the individualized objective functions of specific agents in the vicinity of the robot we control. This online estimation process occurs while interacting with the other agents so that the robot can adapt to each agent individually. For instance, our approach allows for estimating the level of aggressiveness of a specific driver in the surroundings of the autonomous car. 
    
    The online estimation approach we propose is complementary to classical offline IRL methods: With IRL, we can learn a relevant set of objective-function features 
    from real data, as well as a prior distribution over the objective-function parameters. Given these features and a prior on the parameters, we can use LUCIDGames to refine the parameter estimation for a specific agent based on online observation of this agent. 
    Our approach assumes that agents solve a dynamic game and follow Nash equilibrium strategies. This setting models non-cooperating agents that act optimally to satisfy their individual objectives \cite{LeCleach2020a, Fridovich-Keil2020}. 
    We further assume that we have access to a class of objective functions parameterized by a small number of parameters. This could be the desired speed, or driver aggressiveness in the autonomous driving context. 
    
    To estimate these parameters, we adopt the unscented Kalman filtering (UKF) approach. In our case, the key part of this algorithm is the measurement model that maps the objective function parameters to the observation of the surrounding agents' next state. To obtain this mapping, we use ALGAMES, a trajectory optimization solver for dynamic games that handles general nonlinear state and input constraints \cite{LeCleach2020a}.The choice of a derivative-free estimation method (UKF) is justified by the complexity of the measurement model, which includes multiple non-convex constrained optimization problems. Additionally, we design a planner for the robot that is robust to poor estimates of the other agents' objectives. By sampling from the belief over the objective functions of the other agents and computing trajectories corresponding to those samples, we can translate the uncertainty in objective functions into uncertainty in predicted trajectories. Then, ellipsoidal bounds are fitted to the sampled trajectories to form ``safety constraints''; collision constraints that account for objective uncertainty. Importantly, the calculation of these safety constraints reuses samples required by the UKF estimation algorithm. It is, therefore, executed at a negligible additional cost. In a receding-horizon loop, LUCIDGames controls one agent called the ``robot'' and estimates the other agents' objectives at 40 Hz for a 3-player game with a strong level of interaction among the agents. Our primary contributions are as follows:
    \begin{enumerate}
        \item We propose a UKF-based method for a robot to estimate the objective function parameters of non-cooperating agents online, and show convergence of the estimate to the ground-truth parameters.  (Fig. \ref{fig:theta_error}).
        \item We combine the online parameter estimator with a game-theoretic planner.  The combined estimator and planner, called LUCIDGames, runs online at 40Hz in a receding-horizon fashion. 
        \item We include safety constraints within LUCIDGames to impose ellipsoidal collision-avoidance constraints for the robot that reflect the uncertainty in the other agents' future trajectories due to the Bayesian estimate of their parameters.
    \end{enumerate}
    We compare LUCIDGames against game-theoretic and non-game-theoretic baselines. We show that LUCIDGames' trajectory-prediction error rapidly decreases to match the accuracy of the oracle predictor that has access to the ground-truth objectives (Fig. \ref{fig:traj_error}).  Furthermore, we show that LUCIDGames with safety constraints allows for realistic, cautious interactions between a robot and another agent making a unexpected maneuver (Fig.~\ref{fig:ellipse_planning}).

\section{Related Work}
    \subsection{Game-Theoretic Trajectory Optimization}
    Dynamic games have been used as a modeling framework in a wide variety of applications. For example, in autonomous driving \cite{Fridovich-Keil2020, LeCleach2020a}, power system control \cite{Chen2015}, drone and car racing \cite{Spica2018} etc. The solutions of a dynamic game depend on the type of equilibrium selected \cite{Basar1998}. 
    Nash equilibrium  models games without hierarchy  between  players; Each player’s strategy is the best response to the other players’ strategies. 
    Nash equilibrium solutions have been studied extensively \cite{Fridovich-Keil2020, Spica2018, Chen2015}. They seem to capture the game-theoretic interactions observed in some multi-agent non-cooperative problems, e.g. ramp merging. We follow this approach by solving for open-loop Nash equilibrium strategies. However, we intend to relax a key assumption made in previous works by estimating the other agents' objective functions instead of assuming that they are known \emph{a priori} by the robot we control. 
    
    \subsection{Objective Function Estimation}
    Estimating objective functions from historical data is a well investigated problem known as Inverse Optimal Control (IOC) or Inverse Reinforcement Learning (IRL). Typically, with the IRL approach, the objective function is linear in terms of a given set of state features \cite{Ziebart2008}. The goal is to identify a parameter vector that weights these features so that the behavior resulting from this estimated objective matches the observed behavior. While these classical approaches are usually framed in the discrete state and action space setting, they can also be applied to continuous state and action spaces arising in robotics problems \cite{Ratliff2009, Mombaur2010}. However, these works are limited to single-agent problems. 
    
    
    In the multi-agent setting, some IRL approaches formulate the problem by assuming cooperative agents \cite{Hadfield-Menell2016} or competing agents \cite{Waugh2013, Ziebart2010}. These approaches have been demonstrated on discretized state and actions spaces. More recent works consider the multi-agent competitive setting with continuous state and action spaces \cite{Inga2019, Molloy2020}. Their methods have typically been demonstrated on linear-quadratic games with low-dimensional states and control inputs. 
    IOC and IRL-based techniques estimate the objective function's parameters ``offline". Given a set of complete trajectories, they intend to identify one parameter vector that will best fit the data. Our goal is slightly different: As an agent in the game, we would like to perform the estimation ``online", with only knowledge of previous steps, and use our estimate to inform our actions for future time steps. This means that we have access to fewer demonstrations and that our computation time is limited to ensure real-time execution. On the other hand, we assume a low-dimensional parameter space with a coarse prior. In the multi-agent setting, several approaches were proposed to solve this type of online parameter estimation problem.

    \subsection{Online Parameter Estimation}
    We choose the multi-agent autonomous driving problem of highway driving as our running example throughout this paper. We review approaches proposed for online parameter estimation in this context. The Intelligent Driver Model (IDM) \cite{Treiber2000}, is a rule-based lane-following model balancing two objectives: reaching a desired speed and avoiding collision with the preceding car. This model has few tunable parameters and returns the scalar acceleration of the vehicle along a predefined path. Several works used online filtering techniques to estimate these IDM parameters \cite{Hoermann2017, Bhattacharyya2020}. Schultz et al. used particle filtering techniques to estimate the driver's objective including discrete decision variables like turning left or right \cite{Schulz2018}. These works demonstrated that estimating the surrounding drivers objectives helps better predict their future trajectories. However, this gained information was not used to improve the decision making of the cars.
    A recent work estimated the social value orientation (SVO) of the agent the robot is interacting with \cite{Schwarting2019a}.
    However, this formulation still requires knowledge of the desired lane or desired speed of each agent.
    This is precisely the assumption we want to avoid in the present work.  

    \subsection{Data-Driven Trajectory Prediction}
    
    There is a rich literature on applying data-driven approaches to pedestrian or vehicle trajectory predictions. Such approaches usually require a large corpus of data and are trained offline as a general model to suit multiple agents. On the other hand, we require a small amount of data and find parameters online for a specific agent. Data-driven methods can predict a distribution over future trajectories e.g., 
    offline inverse optimal control with online goal inference \cite{Kitani2012}; Conditional Variational Autoencoders (CVAE) \cite{Schmerling2018a} or Generative Adversarial Networks (GAN) \cite{Bhattacharyya2018a}. 
    Our approach maintains a unimodal belief over objective function parameters,\footnote{
        Our approach can easily be extended to multimodal belief representation of objective function parameters using a Gaussian mixture model.} 
    which translates into a distribution over trajectory predictions. A shortcoming of the CVAE-based or GAN-based methods is that they ignore kinodynamic constraints on the predicted trajectories, allowing cars to move sideways, for instance. 
    Incorporating information like drivable area maps which are common for autonomous driving applications \cite{Aurora2019}, could prevent infeasible trajectory predictions \cite{Bansal2018}. Our approach generates dynamically feasible and collision-free predictions. One notable work in this field is Trajectron$++$ \cite{salzmann2020}. It handles kinodynamic constraints and incorporates drivable area maps, as well as the robot's planned trajectory, to inform the prediction. However, contrary to our algorithm, these data-driven methods ignore collision-avoidance constraints between agents and predict trajectories involving collisions as observed by Bhattacharyya \cite{Bhattacharyya2020} with a learning-based method \cite{Bhattacharyya2018a}.

\section{Problem Statement}
    In a multi-player dynamic game, the robot takes its control decisions using LUCIDGames and carries out all the computation required by the algorithm. We assume the other agents are ``ideal'' players in the game. They have access to the ground-truth objective functions of all the players in the game. They take their control decisions by individually solving for a Nash equilibrium strategy based on these true objective functions and execute them in a receding-horizon loop. 
    This assumption is necessary to generate a human driver model that is reactive to the robot's actions and that maintains coupling between planning and trajectory prediction for the robot.
    Other approaches replayed prerecorded driving data to emulate human driving behavior \cite{Hoermann2017, Bhattacharyya2020, Schwarting2019a}, but this method ignores the reactive nature of human drivers to the robots' decisions. Lane-following models, such as IDM \cite{Bouton2019b} fail to capture complex driving strategies like nudging or changing lanes that our model can generate. Moreover, this assumption is required to avoid the complexity of the robot having to "estimate the estimates" of the other agents. Nevertheless, our algorithm shows strong practical performance even when this assumption is violated. All the experiments in this paper are run with the ``ideal'' agents having noisy estimates of the objectives of the surrounding agents in the scene.   
    We further assume that both the robot and the ideal agents plan by computing open-loop Nash equilibrium trajectories and execute these planned trajectories in a receding horizon loop.

    \subsection{Dynamic Game Nash Equilibrium}
    We focus on the discretized dynamic game setting with $N$ time steps and $M$ players ($1$ robot and $M-1$ agents). We denote $x_k \in \mathbb{R}^n$ the joint state of the system, and $u^{\nu}_k \in \mathbb{R}^{m^{\nu}}$ the control input of player $\nu$ at time step $k$. Player $\nu$'s strategy is a control input sequence $U^{\nu} = [({u_1^{\nu}})^T \ldots ({u_{N-1}^{\nu}})^T]^T \in \mathbb{R}^{\bar{m}^\nu}$ where $\bar{m}^{\nu} = (N-1)m^{\nu}$. The robot's strategy is denoted, $U^r$, with $r \in \{1,\ldots,M \}$ and $U^{-r}$ designates the strategies of the $M-1$ other agents in the game.  The state trajectory is defined as $X = [({x_1})^T \ldots ({x_{N}})^T]^T \in \mathbb{R}^{\bar{n}}$ where $\bar{n} = N n$. It stems from executing the control strategies of all the players in the game on a joint dynamical system, 
    \vspace{+0.5mm}
    \begin{align}
        x_{k+1} = f(x_k, u^1_k, \ldots, u^M_k) = f(x_k, u_k),
        \label{eq:discrete_dynamics}
    \end{align}
    with $k$ denoting the time step index. 
    We define the objective function of player $\nu$; $J^{\nu}(X, U^{\nu}): \mathbb{R}^{\bar{n}+\bar{m}^{\nu}} \mapsto \mathbb{R}$. It is a function of its strategy, $U^{\nu}$, and of the state trajectory of the joint system, $X$. The goal of player $\nu$ is to select a strategy, $U^{\nu}$, that will minimize its cost, $J^{\nu}$, while respecting kinodynamic and collision-avoidance constraints. We compactly express these constraints as a set of inequalities $C:\mathbb{R}^{\bar{n}+\bar{m}} \mapsto \mathbb{R}^{n_c}$: 
    \vspace{+0.5mm}
    \begin{mini}[2]
    {X, U^{\nu}}{J^{\nu}(X, U^{\nu}),}{}{}
    \addConstraint{C(X, U) \leq 0}
    \label{pb:gnep}.
    \end{mini}
    Finding a Nash-equilibrium solution to the set of $M$ Problems (\ref{pb:gnep}) is called a generalized Nash equilibrium problem (GNEP) \cite{LeCleach2020a, Facchinei2007}. It consists of finding an open-loop Nash equilibrium control trajectory, i.e. a vector, $\hat{U}$ such that, for all $\nu = 1, \ldots, M$, $\hat{U}^{\nu}$ is a solution to (\ref{pb:gnep}) with the other players' strategies set to $\hat{U}^{-\nu}$. This implies that at a Nash equilibrium point, $\hat{U}$, no player can decrease their objective function by unilaterally modifying their strategy, $U^{\nu}$, to any other feasible point.
    Solving this GNEP can be done efficiently with a dynamic game solver such as ALGAMES \cite{LeCleach2020a}. We will consider it as an algorithmic module that takes as inputs the initial state of the system and the objective functions, $J^1, \ldots, J^M$, and returns an open-loop trajectory of the joint system comprising the robot and the ideal agents.


    \subsection{Objective Function Parameterization}
    As is typically the case in the IRL and IOC literature \cite{Ziebart2008, Mombaur2010, Inga2019}, we assume that the objective function of player $\nu$ can be expressed as a linear combination of features, $\phi$, extracted from the state and control trajectories of this player,
    \begin{align}
        J^{\nu}(X, U^{\nu}) = \phi(X, U^{\nu})^T \theta^{\nu}.
        \label{eq:linear_parameterization}
    \end{align}
    While restrictive, this parameterization encompasses many common objective functions like linear and quadratic costs. 
    The UKF estimates the weight vector $\theta^{\nu}$ of all the agents in the game. We denote by $\theta \in \mathbb{R}^q$ the concatenation of the vectors $\theta^{\nu}$ that the robot has to estimate,
    \begin{align}
        \theta = [{\theta^{1}}^T \ldots \:\:\: {\theta^{r-1}}^T, {\theta^{r+1}}^T  \ldots \:\:\: {\theta^{M}}^T]^T \: \in \: \mathbb{R}^q.
    \end{align}

\section{Unscented Kalman Filtering Formulation}   


    We propose an algorithm that allows the robot to estimate the objective functions' parameter $\theta$ and to exploit this estimation to predict the other agents' behaviors and make decisions for itself. We represent the belief over the parameter $\theta$ as a Gaussian distribution and we sample sigma-points from it. Each sigma-point is a guess over the parameter $\theta$. Given the current state of the system, $x$, we can form a GNEP for each sigma-point. By solving these GNEPs, we obtain a set of predicted trajectories for the system. When we receive a new measurement of the state $x$, we compare it to the trajectories we predicted earlier. The Gaussian belief over $\theta$ is updated with the typical Unscented Kalman Filter (UKF) \cite{Wan2000} update rules, so that the sigma-points that had better prediction performance are now more likely.  
    
    The UKF framework requires two pieces: the process model, and the measurement model.  In a typical filtering context, these are obvious.  However, in our problem these are more subtle.  Specifically, the quantity we estimate with the filter is $\theta$.  We assume this quantity evolves according to a random walk, so the process model for the UKF is the identity map plus Gaussian white process noise.  The crucial part of our algorithm is the measurement model.  In our case, the measured quantity available to the robot (with noise) is the system state, $x_t$, at the current time.  Hence, the measurement model is the map relating the parameter vector $\theta$ to the system state $x_t$.  This function is itself the solution of the dynamic game.  Therefore, our UKF requires the solution of the dynamic game for each sigma-point of $\theta$ at each time step.  The dynamics and measurement models, as well as the steps in our UKF estimator are described in detail below.

    

\enlargethispage{-1.9mm}
    \subsection{Process and Measurement Model}
    Our estimator is executed in a receding-horizon loop. At each time step $t$, the robot updates its Gaussian belief over the vector $\theta$, which is parameterized by its mean, $\mu_t$, and covariance matrix, $\Sigma_t$. We assume that the ground-truth parameter $\theta$ is a random walk with relatively small process noise covariance, which means that the agents' objectives are nearly constant, but may change slightly over the course of the estimation. This is a reasonable assumption as, for many robotics applications, an agent's objective corresponds to its long-term goal and thus varies over time scales far larger than the estimator's update period. The process model corresponds to an additive white Gaussian noise and is defined as follows,
    \begin{align}
        \label{eq:process_model}
        &\theta_{t+1} = \theta_t + \delta_t, 
        &\delta_t \sim \mathcal{N}(0, Q_t).
    \end{align}

    \vspace{-1mm}
    We construct a measurement model, $g(\cdot,\cdot)$, that maps the parameter $\theta$ and the observed previous state $x_{t-1}$ to the current state $x_t$ that we observe\footnote{A direction for future work could be to consider the case where the robot has a nonlinear or partial observation of the state.}:
    \begin{align}
        \label{eq:meas_model}
        x_{t} &= g(\theta, x_{t-1}) + \epsilon_t, \\
        \label{eq:meas_noise}
        \epsilon_t &\sim \mathcal{N}(0, R).
    \end{align}
    This nonlinear function, $g(\cdot,\cdot)$, encapsulates the decision making process of the agents and the propagation of the system's dynamics as detailed in Algorithm \ref{al:controller}. 
    

    \subsection{UKF Algorithm}
    The estimator propagates the belief over the vector $\theta$ in time. The procedure that updates this belief is described in Algorithm \ref{al:estimator}. Lines 2 and 3 correspond to the prediction step, which exploits the process model. Line 4 samples sigma-points from a Gaussian distribution over the vector $\theta$ (Eq. \ref{eq:sigma_1}-\ref{eq:sigma_7}),
    \vspace{-3mm}
    
    \begin{align}
        \label{eq:sigma_1}
        \lambda &= \alpha^2 (q+\kappa) - q, \\
        \label{eq:sigma_2}
        \Theta^{(0)} &= \mu, \\
        \label{eq:sigma_3}
        \Theta^{(i)} &= \mu + (\sqrt{(q+\lambda)\Sigma})_i, 
        &&\hspace{-3mm} \forall i \in \{1,\ldots,q\}\\
        \label{eq:sigma_4}
        \Theta^{(i)} &= \mu - (\sqrt{(q+\lambda)\Sigma})_{i-q},
        &&\hspace{-3mm} \forall i \in \{q+1,\ldots,2q\}\\
        \label{eq:sigma_5}
        M^{(0)} &= \lambda / (q+\lambda), \\
        \label{eq:sigma_6}
        C^{(0)} &= \lambda / (q+\lambda) + (1 - \alpha^2 + \beta), \\
        \label{eq:sigma_7}
        M^{(i)} &= C^{(i)} = \lambda / \big( 2(q+\lambda) \big). 
        &&\hspace{-3mm} \forall i \in \{1,\dots,2q\}
    \end{align}
    
    We follow a classical sampling scheme that relies on parameters, $\alpha$, $\beta$, $\kappa$, controlling the spread of the sigma-points and encoding prior knowledge about the distribution. Wan et al. provide a detailed interpretation for these parameters \cite{Wan2000}. Line 5 applies the measurement model to the sampled sigma-points, $\Theta$, which are guesses over the vector $\theta$. Specifically, for each sigma-point, we solve the dynamic games that the ideal agents and the robot encountered at the previous time step. Then, we propagate the system's dynamics for one time step to obtain $\chi$, a set of predictions over $x_t$, the current state of the system. Lines 6 to 9 compute the Kalman gain, $K$, and measurement prediction $\bar{x}_t$. Finally, the update step is executed in lines 10 and 11.

    
    \setlength{\textfloatsep}{0pt} 
    \begin{algorithm}[t]
    \caption{Parameter estimation module. }\label{al:estimator}
    \begin{algorithmic}[1]
    \Procedure{Estimator}{$x_{t-1}$, $x_{t}$, $\mu_{t-1}$, $\mu_{t}$, $\Sigma_{t}$}
        \State \hspace{-3mm} $\bar{\mu}_{t+1} \gets \mu_{t} $
        \State \hspace{-3mm} $\bar{\Sigma}_{t+1} \gets \Sigma_{t} + Q_t $
        \State \hspace{-3mm} $\Theta, M, C \gets $ \Call{SigmaPoints}{$\bar{\mu}_{t+1}, \bar{\Sigma}_{t+1}$} \Comment{Eq. \ref{eq:sigma_1}-\ref{eq:sigma_7}}
        \State \hspace{-3mm} $\chi^{(i)} \gets$ \Call{Controller}{$x_{t-1}, \mu_{t-1}, \Theta^{(i)}$} 
        \hfill $ \forall i$
        \State \hspace{-3mm} ${\bar{x}}_{t} \gets \sum_{i} M^{(i)} {\chi}^{(i)}$
        \State \hspace{-3mm} $P \gets \sum_{i} C^{(i)} [{\chi}^{(i)} - \bar{x}_t] [{\chi}^{(i)} - \bar{x}_t]^T$
        \State \hspace{-3mm} $S \gets \sum_{i} C^{(i)} [\Theta^{(i)} - \bar{\mu}_{t+1}] [{\chi}^{(i)} - \bar{x}_t]^T$
        \State \hspace{-3mm} $K \gets S {P}^{-1}$
        \State \hspace{-3mm} $\mu_{t+1} \gets \bar{\mu}_{t+1} + K(\chi - \bar{x}_t)$
        \State \hspace{-3mm} $\Sigma_{t+1} \gets \bar{\Sigma}_{t+1} - K P K^T$
    \State \hspace{-3mm} \textbf{return} ${\mu_{t+1}, \Sigma_{t+1}}$
    \EndProcedure
    \end{algorithmic}
    \end{algorithm} 

    
    \setlength{\textfloatsep}{0pt} 
    \begin{algorithm}[t]
    \caption{Decision making process of the agents.}\label{al:controller}
    \begin{algorithmic}[1]
    \Procedure{Controller}{$x_t$, $\mu_t$, $\theta$}
        \State \hspace{-3mm} $U^r_t$ $\gets$ \Call{ALGAMES}{$x_t$, $\mu_t$} \Comment{Robot's plan}
        \State \hspace{-3mm} $U^{-r}_t$ $\gets$ \Call{ALGAMES}{$x_t$, $\theta$} \Comment{Ideal agents' plans}
        \State \hspace{-3mm} $U_t \gets [{U^r_t}^T, {U^{-r}_t}^T]^T$ 
        \State \hspace{-3mm} $x_{t+1}$ $\gets$ \Call{Dynamics}{$x_t$, $U_t$} \Comment{Equation \ref{eq:discrete_dynamics}}
        \State \hspace{-3mm} \textbf{return} ${x_{t+1}}$
    \EndProcedure
    \end{algorithmic}
    \end{algorithm}
\addtolength{\topmargin}{+1.9mm}
\enlargethispage{-1.9mm}
	\subsection{LUCIDGames: Combining Parameter Estimation and Planning}
    
    
    LUCIDGames exploits the information gained via the estimator to inform the decision making of the robot. It jointly plans for itself and predicts the other agents' trajectories. At time step $t$, the robot solves the GNEP using the current state of the system $x_t$ and its mean estimate $\mu_t$ over $\theta$. We obtain the next state $x_{t+1}$ by propagating forward the open-loop plans of both the robot and the ideal agents as detailed in Algorithm \ref{al:controller}. The joint estimation and control procedure is detailed in Algorithm \ref{al:lucidgames}. 

    \begin{algorithm}[t]
    \caption{Combined estimator and planning module.}\label{al:lucidgames}
    \begin{algorithmic}[1]
    \Procedure{LUCIDGames}{$x_{t-1}$, $x_{t}$, $\mu_{t-1}$, $\mu_{t}$, $\Sigma_{t}$}
        \State \hspace{-3mm} \textbf{for} $t = 1, 2, \ldots$ \textbf{do} 
            \State \hspace{-0mm} $x_{t+1} \gets $ 
            \Call{Controller}{$x_t$, $\mu_t$, $\theta$}
            \State \hspace{-0mm} $\mu_{t+1}, \Sigma_{t+1} \gets$ \Call{Estimator}{$x_{t-1}, x_t, \mu_{t-1}, \mu_t, \Sigma_t$} 
        \State \hspace{-3mm} \textbf{return} ${x_{t+1}, \mu_{t+1}, \Sigma_{t+1}}$
    \EndProcedure
    \end{algorithmic}
    \end{algorithm} 

    \begin{figure}[t]
        \centering
        \includegraphics[width=8.60cm]{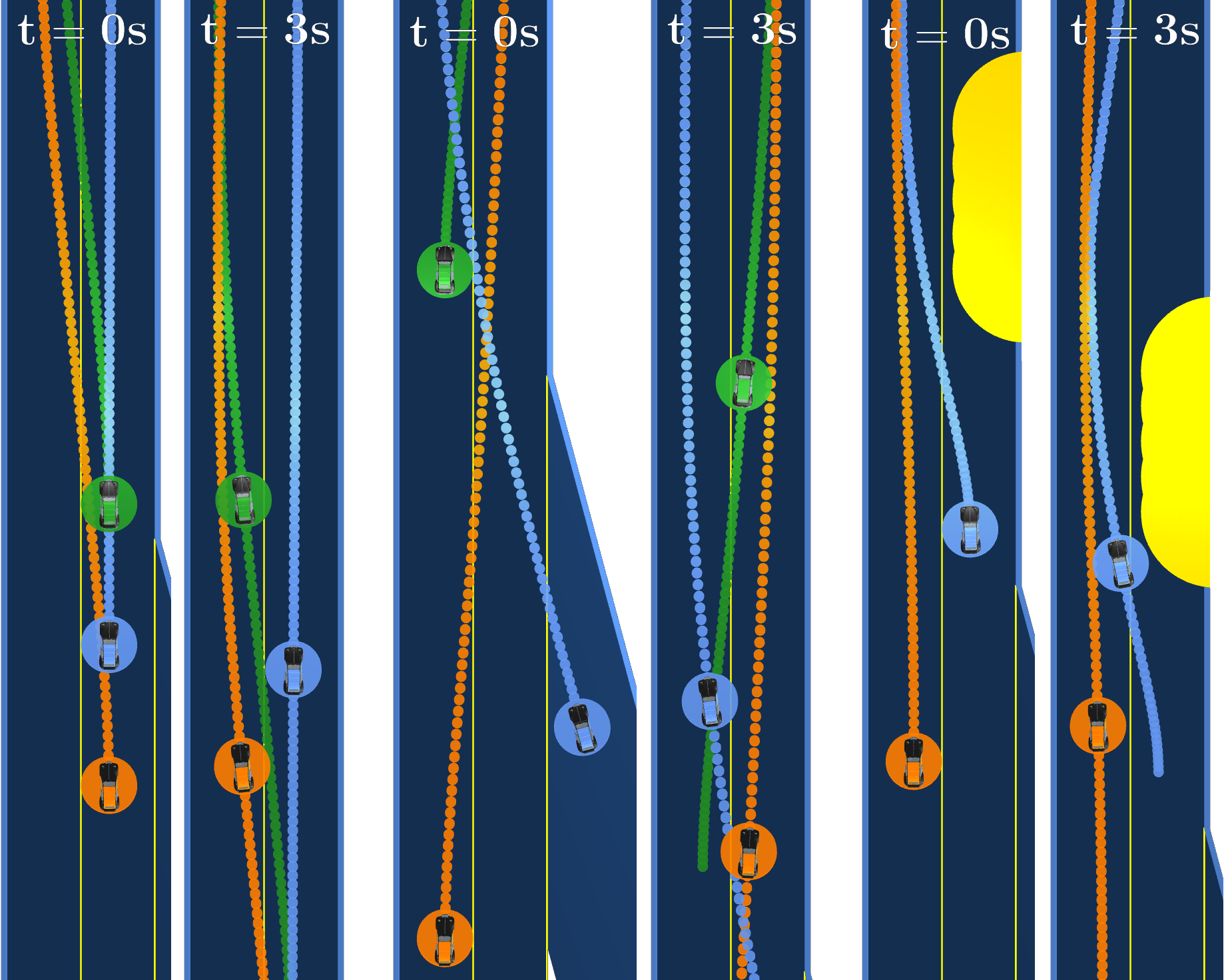}\hfill
        \caption{From left to right, we represent three highway driving scenarios, an overtaking maneuver, a ramp merging, and an obstacle avoidance maneuver. For each scenario we represent the system at $t = 0$s and $t = 3$s.}
        \label{fig:scenarios}
    \vspace{-5mm}
    \end{figure}
    \setlength{\textfloatsep}{\textfloatsepsave}

\enlargethispage{-1.9mm}
\vspace{-0.5mm}
\section{Simulations: Design and Setup}
    We apply our algorithm to highway autonomous driving problems involving a high level of interactions between agents. Specifically, we test LUCIDGames in three driving scenarios exhibiting maneuvers such as overtaking, ramp merging and obstacle avoidance (Figure \ref{fig:scenarios}). We assume the robot follows the LUCIDGames algorithm for its decision making and estimation. The other vehicles are modeled as ideal agents solving the dynamic game with knowledge of the true parameters. 

    \subsubsection{Problem Constraints}
    We consider a  unicycle model for the dynamics of each vehicle. The state, $x_k$, contains a 2D position, a heading angle and a scalar velocity for each vehicle. The control input, $u^{\nu}_k$, consists of an angular velocity and a scalar acceleration. 
    Additionally, we model the collision avoidance zone of each vehicle as a disk, preventing collision between vehicles and with the boundaries of the road.
    
\enlargethispage{-1.9mm}
    \subsubsection{Objective Function}
    We select a quadratic objective function incentivizing the agents to reach a desired state, $x_f$, while limiting control inputs. On top of this objective function, we add a quadratic penalty on being close to other vehicles,
    \begin{align}
        J^{\nu}&(X, U^{\nu}) = \sum_{k=1}^{N-1} \frac{1}{2}(x_k - x_f)^T Q (x_k - x_f) + \frac{1}{2}{u^{\nu}_k}^T R u^{\nu}_k + \nonumber \\[-6pt]
        & \frac{1}{2} (x_N - x_f)^T Q_f (x_N - x_f) + \nonumber \\[-6pt]
        &\:\:\:\:\sum_{k=1}^{N} \sum_{\mu \neq \nu} \gamma^{\nu} \bigg( \max{\big(0, || p_k^{\nu} - p_k^{\mu} ||_2 - \eta (r^{\nu} + r^{\mu})\big)} \bigg)^2.
        \label{eq:lqr_cost}
    \end{align}
    For agent $\nu$, $p_k^{\nu}$ and $r^{\nu}$ designate its 2D position at time step $k$ and collision avoidance radius. $\gamma^{\nu}$ and $\eta$ are scalar collision avoidance cost parameters encoding the magnitude of the cost and the distance at which this cost is ``activated''. 
    
    In this work, we estimate a reduced number of objective function parameters. We choose 3 parameters with intuitive interpretations. Two of them are elements of the desired state, $x_f^{\nu}$. They correspond to the desired speed and desired lateral position on the roadway (i.e. desired lane) of the vehicle. The last one is $\gamma^{\nu}$, which encodes the ``aggressiveness'' of the driver. Indeed, a large value for $\gamma^{\nu}$ will penalize a vehicle driving too close to other vehicles, which will lead to less aggressive behavior. We remark that this parameterization is consistent with an objective function expressed as a linear combination of features, as in Equation \ref{eq:linear_parameterization}. Therefore, it would be possible to use an IRL algorithm trained on real driving data to provide a prior on these parameters. 

\section{Simulation Results}
    To assess the merits of LUCIDGames, we test it on highway driving scenarios as shown on Figure \ref{fig:scenarios}. We first assess the tractability and scalability of the approach for an increasing number of agents. Then, we perform an ablation study by removing the two main components of LUCIDGames: the online estimation and the game-theoretic reasoning. The goal is to investigate how each of these components affect the performance of LUCIDGames. This is also a way to compare LUCIDGames to related approaches proposed in the literature. Indeed, several works applied dynamic game solvers in a receding-horizon loop to autonomous driving problems without resorting to online estimation \cite{LeCleach2020a, Fridovich-Keil2020}. Bhattacharyya et al. \cite{Bhattacharyya2020} used highway driving datasets to compare the trajectory prediction performance of rule-based and black-box driver models. The set of evaluated methods included among others: constant velocity prediction, Generative Adversarial Imitation Learning (GAIL) \cite{Bhattacharyya2018a} and a particle filter estimating the parameters of the intelligent driver model (IDM) online. On the trajectory prediction task, the constant velocity baseline was the best performing method. Thus, we choose to compare our method to this non-game-theoretic baseline. 
    
    
\enlargethispage{-1.9mm}
    \subsection{Tractability}
    We run LUCIDGames in a receding horizon-loop using a coarse prior on the vector $\theta$. In practice, the initial belief is a Gaussian parameterized by its mean and variance:
    \begin{align}
        &\mu_0 = \mathbf{1}, &\Sigma_0 = v_0 I_q.
    \end{align}
    Where $v_0$ is a large initial variance on each parameter (typically $v_0 = 25$ in our experiments); and $I_q$ is the identity matrix. We run LUCIDGames on the ramp-merging scenario involving 2 to 4 agents and we compile the timing results in Table \ref{fig:mpc_table}. We demonstrate the tractability of the algorithm for complex autonomous driving scenarios, and we show real-time performance of the estimator for three agents (40Hz) and up to four agents ($10$Hz). 
    In practice, we trivially parallelize the implementation of LUCIDGames: For each sigma-point, $\Theta_t^{(i)}$, the algorithm requires the solution of a dynamic game (Algorithm \ref{al:estimator}, line 5). We solve these dynamic games simultaneously, in parallel, by distributing them on a multi-core processor. The number of dynamic games to solve in parallel scales like the number of sigma-points, which is linear in terms of the number of agents $M$. Each individual dynamic game has a computational complexity of $O(M^3)$. In this work, all the experiments have been executed on a 16-core processor (AMD Ryzen 2950X). 
    
    \begin{figure}[t]
        \vspace{+1.8mm}
        \begin{center}
        \begin{tabular}{|c c c c|} 
            \hline
            $\#$ of Players     & Freq. in Hz & $\mathbb{E}[\delta t]$ in ms & $\mathbb{\sigma}[\delta t]$ in ms \\ [0.5ex] 
            \hline
            2 & 132  &  7.55 &  16.7 \\ 
            3 & 38.0 &  26.3 &  37.7 \\
            4 & 11.4 &  87.3 &  94.0 \\
            \hline
        \end{tabular}
        \vspace{+0.3cm}
        \caption{Running the MPC implementation of LUCIDGames 50 times on a ramp-merging scenario with 2,3 and 4 players, we obtain the mean update frequency of the MPC as well as the mean and standard deviation of $\delta t$, the time required to update the MPC plan and the belief over the other players' objective functions.}
        \vspace{-0.6cm}
        \label{fig:mpc_table}
        \end{center}
    \vspace{-1mm}
    \end{figure}

    \begin{figure}[t]
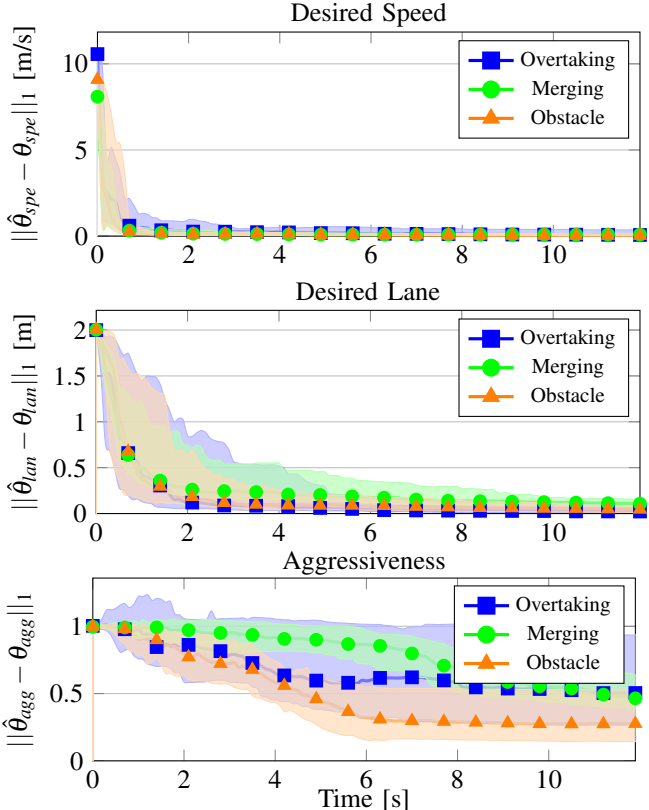

        \centering
        \includegraphics[width=8.60cm, height=3.7cm]{tikz/parameter_estimation_full/parameter_estimation_Theta_speed_error_temp_10.tikz}
        \vspace{-1mm}
        \includegraphics[width=8.60cm, height=3.7cm]{tikz/parameter_estimation_full/parameter_estimation_Theta_lane_error_temp_10.tikz}
        \vspace{-1mm}
        \includegraphics[width=8.60cm, height=3.7cm]{tikz/parameter_estimation_full/parameter_estimation_Theta_agg_error_temp_10.tikz}
        \vspace{-1mm}
        \caption{LUCIDGames reduces the estimation error on the desired speed parameter by a factor of 100 within 12 seconds of interaction (top plot). The error on the desired lane is divided by 20 (middle plot) and the error on the aggressiveness parameter is halved (bottom plot). The markers indicate the median error computed over 50 simulations. The faded color areas correspond to the interval between the 1$^{st}$ and 3$^{rd}$ quantile.}
        \label{fig:theta_error}
        \vspace{-6mm}
    \end{figure}
    
\addtolength{\topmargin}{-1.9mm}
    \subsection{Parameter Estimation}
    We assess the ability of LUCIDGames to correctly estimate the ground-truth objectives of the other agents with only a few seconds of driving interaction. We test LUCIDGames on three scenarios: highway overtaking, ramp merging and obstacle avoidance. We compute the relative error between the ground-truth parameter $\theta$ and the mean of the Gaussian belief $\mu_t$ along the 12-second MPC simulation. We perform a Monte-Carlo analysis of the algorithm by sampling the initial state of the system as well as the objective parameter $\theta$. The aggregated results from 50 samples are presented in Figure \ref{fig:theta_error}. We observe a significant decrease in the relative error between the ground-truth parameter and the mean of the belief.

    \subsection{Trajectory Prediction}

    \begin{figure}[t]
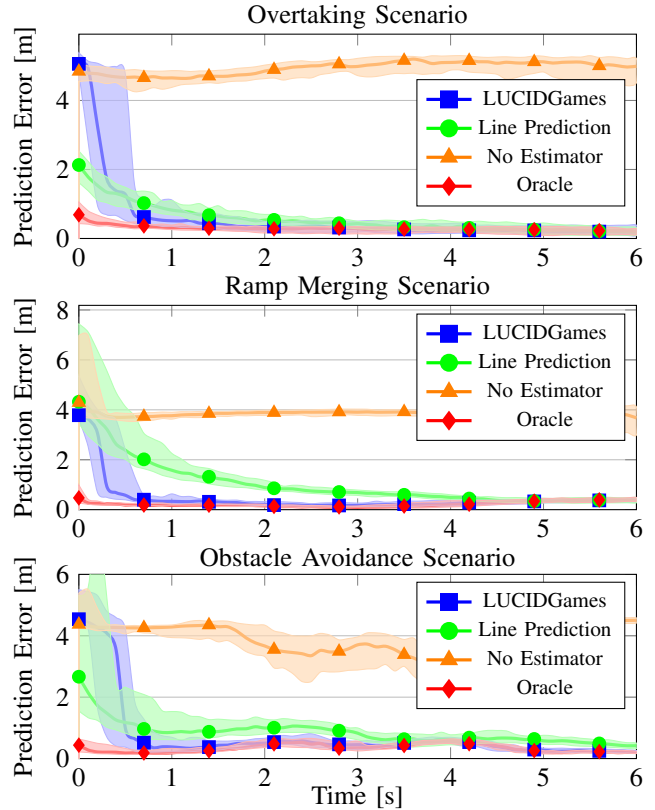

    \vspace{+0.8mm}
        \centering
        \includegraphics[width=8.60cm, height=3.7cm]{tikz/trajectory_prediction_full/trajectory_prediction_Straight_planning_error_temp_10.tikz}\hfill
        \vspace{-1mm}
        \includegraphics[width=8.60cm, height=3.7cm]{tikz/trajectory_prediction_full/trajectory_prediction_Merging_planning_error_temp_10.tikz}\hfill
        \vspace{-1mm}
        \includegraphics[width=8.60cm, height=3.7cm]{tikz/trajectory_prediction_full/trajectory_prediction_Obstacle_planning_error_temp_10.tikz}\hfill
        \vspace{-1mm}
        \caption{We present the trajectory prediction error obtained by the robot on 3 scenarios for 4 algorithms: LUCIDGames; a non-game-theoretic baseline using straight line predictions; a game-theoretic solver that does not estimate the other agents' objective functions and finally, an oracle having access to the ground-truth objective functions of the other agents. LUCIDGames starts with a large prediction error and quickly estimates the other agents' objective functions to outperform the baselines and reach error levels comparable to those of the oracle. We represent the medians, computed over 50 simulations, of the prediction error measuring the $\ell 2$-distance between the 3-second trajectory predictions and the ground-truth trajectory. The faded color areas correspond to the interval between the 1$^{st}$ and 3$^{rd}$ quantile. }
        \label{fig:traj_error}
        \vspace{-7mm}
    \end{figure}

    Additionally, we show that LUCIDGames allows the robot to better predict the trajectory of the other agents. We use the same Monte-Carlo analysis as described above on three driving scenario (Fig. \ref{fig:traj_error}). First, we observe that LUCIDGames initialized with a coarse prior starts with a large prediction error; around 4 m in all scenarios. However, it converges to a prediction error very much comparable to the one obtained with the oracle in about 1 second. This illustrates the ability of the robot using LUCIDGames to quickly improve its predictions about the surrounding agents by gathering information about them. The error obtained by keeping the coarse prior remains high and fairly constant during the simulation. The one obtained using LUCIDGames is an order of magnitude lower, in comparison, by the end of the simulation. 
    
    Second, we compare LUCIDGames to a non-game-theoretic baseline on trajectory prediction error. This baseline predicts the trajectories of the agents surrounding the robot by propagating straight-line and constant-speed trajectories for each agent. These predicted trajectories are of the same duration (3 seconds) as the open-loop predictions made by LUCIDGames. We choose to compare our approach to this baseline to assess the impact of including game-theoretic reasoning on the trajectory prediction performance.  
    This line-prediction baseline may seem very coarse. However, in the context of highway driving, straight line trajectories are very pertinent for short (3 seconds) horizon predictions. In practice, we use a straight highway environment for our simulations (Figure \ref{fig:scenarios}). As the roadway is not curved, the only causes of trajectory curvature are lane changes, nudging and merging maneuvers. LUCIDGames is able to outperform the baseline by capturing these natural driving behaviors that go beyond lane following.
    
    For the overtaking scenario (Figure \ref{fig:traj_error}), LUCIDGames starts off with a large prediction error but quickly converges to prediction error lower than the line-prediction baseline. However, the performance gap is small confirming that the line prediction baseline is a suitable model for short horizon prediction in typical highway driving. 
    
    On the other hand, for more complex driving scenarios like ramp merging, the gap between LUCIDGames and the line prediction technique is significant. We observe that this gap is the highest after 1 second when LUCIDGames has successfully converged. The gap consistently decreases afterwards as the system converges to a steady state where all the vehicles drive in straight lines following their desired lanes. 
    
    Similarly, for the collision-avoidance scenario, the prediction error obtained using LUCIDGames is around half that obtained using the line-prediction baseline after the parameter estimation has converged (1 second). 
    We observe that the line prediction baseline almost matches LUCIDGames' when the vehicles are constrained to drive on a narrower roadway ($t \in [3,4]$s). Finally, after the obstacle is passed ($t \in [4,6]$s), the performance gap increases in favor of LUCIDGames. Indeed, it is able to predict that vehicles are going to return to their desired lanes after avoiding the obstacle.  

    \subsection{Safe Trajectory Planning}
    We implement a robust trajectory planning scheme for the robot that accounts for uncertainty in the objective of the other agents by enforcing ``safety constraints.'' With LUCIDGames, we maintain a Gaussian belief over the other agents' objectives. We thus quantify the uncertainty of our current objective function estimates. Taking into account such uncertainty can be instrumental in preventing the robot from making unsafe decisions. For instance, an autonomous vehicle should act cautiously when overtaking an agent for which it has an uncertain estimate of its desired speed and desired lane. 
    
    For many multi-robot systems, safety is ensured by avoiding collisions with other agents. Thus, we encode safe decision making for the robot by ensuring its decisions are robust to misestimation of the objective functions. First, the robot computes the ``safety constraints,'' which are inflated collision avoidance constraints around other agents by fitting ellipses around the trajectories sampled by the UKF (e.g., in the 95\%-confidence ellipse). These safety constraints can be seen as approximate chance constraints that can be efficiently computed.  
    Then, the robot solves the dynamic game corresponding to the mean of the belief over $\theta$, with the safety constraints. In practice, when the uncertainty about $\theta$ is large, the sampled sigma-points and their corresponding trajectories are scattered and generate a large collision avoidance zone. The top roadway visualization in Figure \ref{fig:ellipse_planning} illustrates this situation. These safety constraints can be seen as a lifting of the uncertainty in the low-dimensional space of objective parameters onto the high-dimensional space of predicted trajectories. The ``keep-out'' zone is cone-shaped in free space as expected but shrinks down when the roadway narrows or when the sampled trajectories concur towards the same position. 

    We showcase the driving strategy emerging from this robust planning scheme in Figure \ref{fig:ellipse_planning}. The human driver and the robot are confronted with an obstacle. Using LUCIDGames, the robot infers the human's intent to change lanes (to avoid the obstacle), and negotiates, through the game theoretic planner, whether to yield to the human, or to let the human yield. In phase 1, the robot has a large initial uncertainty about the objective of the human-driven vehicle (blue). Indeed, the set of sampled trajectories contains both predictions where the human cuts in front of the robot, and ones in which the human yields to let the robot go first. Thus, the robot slows down to comply with the safety constraints which covers both hypotheses. In phase 2, the robot has correctly estimated the human's intent to yield to the robot, to change lanes after the robot passes. Since the robot's estimate of the human's objective is more certain, the collision avoidance zone generated by the safety constraints shrinks. This allows the robot to plan an overtaking maneuver and regain speed. In phase 3, the robot safely proceeds in its own lane cruising at its desired velocity, while the human changes lanes behind the robot to avoid the obstacle. On the other hand, LUCIDGames without these safety constraints does not slow down to account for the initial uncertainty. The same is true for the oracle and the straight line prediction baseline. We observe the same behaviors on the scenario where the robot yields to the human.  

    \subsection{Results Discussion}
    \subsubsection{Tractability}
    We demonstrate the tractability of the algorithm for up to four agents with an update rate of $10$Hz. However, it is important to notice that the robot's controller and its estimator can be run at different rates. This would allow for a fast update of the robot's plan and a slower update of its estimation of the objective functions of the other agents. 
    
    \subsubsection{Parameter Estimation}
    The good estimation performance of LUCIDGames was assessed through a Monte Carlo analysis. However, we have observed challenging scenarios demonstrating the complexity of the objective estimation task. For instance, if all the agents are far from each other, none of the collision avoidance penalties (Eq. \ref{eq:lqr_cost}) are ``active''. In such a situation, it is impossible to estimate the aggressiveness parameter that scales the cost of these collision avoidance penalties.
    Nevertheless, we argue that this observability issue is not crucial in practice. Indeed, as long as the agents remain far from each other the aggressiveness parameter will not affect the trajectory prediction of the robot. Conversely, as soon as the agents get closer to each other, the aggressiveness parameter matters but the observability issue disappears. 
    
    \subsubsection{Trajectory Prediction}
    The human-driven vehicles in these simulations are modeled as agents solving the ground-truth dynamic game for a Nash equilibrium strategy in a receding horizon loop.  We notice in Figure \ref{fig:traj_error} that even when the robot has access to the ground-truth objective functions its trajectory prediction error is not null. This is due the noise added to the dynamics as well as the discrepancy between the open-loop plan predictions and the actual trajectory stemming from executing open-loop strategies in a receding horizon loop. We observe in Figure \ref{fig:traj_error} that LUCIDGames consistently converges to error levels closely matching the ones of the agent having access to the ground-truth objective functions.

\section{Conclusion}
    We have presented LUCIDGames, a game theoretic planning framework that includes the solution of an inverse optimal control problem online to estimate the objective function parameters of other agents. We demonstrate that this algorithm is fast enough to run online in a receding-horizon loop, and is effective in planning for an autonomous vehicle to negotiate complex driving scenarios while interacting with other vehicles. We showed that this method outperforms two benchmark planning algorithms, one assuming straight-line predictions for other agents, and one incorporating game-theoretic planning, but without online parameter estimation of other agents' objective functions. 
    
    We envision several promising directions for future work: In this work, the set of objective function parameters has been designed with ``expert'' knowledge of the problem at hand, so that they encompass a large diversity of driving behaviors while remaining low dimensional. However, one could envision these parameters and associated features being identified via a data-driven approach. The overall approach of estimating online a reduced set of parameters to better predict the behavior of the system is appealing.
    Indeed, in this framework, the dynamic game solver lifts the low dimensional space of objective function parameters (order $10^1$) into the high dimensional space of predicted trajectories (order $10^2$ - $10^3$). This lifting or ``generative model'' natively embeds safety requirements by generating dynamically feasible trajectories respecting collision avoidance constraints. It also accounts for the fact that agents tend to act optimally with respect to some objective functions. Finally, through its game-theoretic nature, it captures the reactive nature of the agents surrounding the robot in autonomous driving scenarios, where negotiation between players is a crucial feature.
    \vspace{-1mm}

{\setstretch{0.95}
\bibliographystyle{ieeetr}
\bibliography{LUCIDGames}
}
\end{document}